\documentclass{article} % For LaTeX2e
\usepackage{nips14submit_e,times}
\usepackage{amsmath,amsthm,verbatim,amssymb,amsfonts,amscd, graphicx}
\usepackage{algorithm}
\usepackage{url}
\usepackage{graphics}
\usepackage{caption}
\usepackage{subfig}
\usepackage[numbers]{natbib}
\graphicspath{ {./} }

\title{Learning Machines Implemented on Non-Deterministic Hardware}

\author{
Suyog Gupta \\
\And
Vikas Sindhwani \\
\And
Kailash Gopalakrishnan \\
\AND
\\
IBM T. J. Watson Research Center\\
Yorktown Heights, NY 10598\\
\texttt{\{suyog,vsindhw,kailash\}@us.ibm.com} \\
}
\nipsfinalcopy % Uncomment for camera-ready version

\begin{document}

\maketitle

\begin{abstract}
This paper highlights new opportunities for designing large-scale machine learning systems as a consequence of blurring traditional boundaries that have allowed algorithm designers and application-level practitioners to stay -- for the most part -- oblivious to the details of the underlying hardware-level implementations. The hardware/software co-design methodology advocated here hinges on the deployment of compute-intensive machine learning kernels onto compute platforms that trade-off determinism in the computation for improvement in speed and/or energy efficiency. To achieve this, we revisit digital stochastic circuits for approximating matrix computations that are ubiquitous in machine learning algorithms. Theoretical and empirical evaluation is undertaken to assess the impact of the hardware-induced computational noise on algorithm performance. As a proof-of-concept, a  stochastic hardware simulator is employed for training deep neural networks for image recognition problems.
\end{abstract}
\section{Introduction}
Applications that automate the process of extracting meaningful insights from an ever-increasing trove of user and sensor-generated data have emerged as one of the dominant consumers of computing resources. The natural error-resilience of a large suite of learning algorithms enabling such applications is well-documented, setting them apart from more traditional workloads that typically require high precision computation and number representations with high dynamic range. The strategy of embracing errors during computation is in fact a binding theme across several disciplines that impact large-scale machine learning. It is well appreciated that in the presence of statistical approximation and estimation errors, high-precision computation in the context of learning is rather unnecessary~\cite{BottouBousquet}. Consequently, stochastic optimization techniques~\cite{KushnerLin} and randomized numerical linear algebra~\cite{Mahoney} are becoming critical components of the lower layers of an optimized machine learning software stack, blurring the computation-statistics interface~\cite{Jordan}. Yet, machine learning applications continue to be deployed on general purpose computing platforms that have been designed to cater to the needs of the traditional workloads, incurring high, and often unnecessary penalty in terms of degradation in the overall system performance.

The motivation for this paper stems from the idea that the learning algorithm's intrinsic robustness to noise may be leveraged to relax certain constraints on the underlying hardware. In the proposed model, the compute engines executing the algorithm perform approximate computations, introducing non-deterministic errors in the process. It is reasonable to expect that this loss in accuracy is accompanied by a corresponding increase in speed and/or energy-efficiency per computation. For instance, reduced precision, fixed point units are typically faster and often consume far less hardware resources and power than floating point engines. If exposing the hardware-generated noise to the algorithm does not result in degradation in terms of a pre-defined measure-of-quality metric, this hardware-software co-design scheme can prove to be a viable approach for optimizing the system performance.

Provoking a discussion along these lines is especially timely, given the increasing likelihood of the demise of Moore's law/Dennard scaling and the resulting collapse of the conventional model of processor design that owes much of its success to the sustainability of transistor area and performance scaling. Our approach involves identifying noise-tolerant kernels that dominate the algorithm run time and offloading the execution of these kernels onto a dedicated hardware accelerator that performs approximate computations while interacting closely with the host processor. As it will be elucidated in the sections to follow, invoking approximations at the compute level adds another dimension (and complexity) in the hardware design space and to truly benefit from such an approach entails careful engineering of several closely coupled aspects of system design. This includes definition of the accelerator microarchitecture and optimizing the host-accelerator interface which involves non-trivial optimization in the design subspace defined by the requirements on performance, accuracy, energy consumption and implementation costs.~It is also equally important to preserve the programming model so that these hardware benefits can be readily absorbed at the application-level without incurring additional software development costs.~Since hardware design is typically beleaguered with substantial engineering costs and longer development time than software, it is not only prudent but also necessary to firmly establish the feasibility of the approximate computing techniques for a given set of target applications. Also, care should be taken to avoid a common pitfall in application-specific integrated circuit design that results in the usefulness of the hardware solution to be limited to only a niche set of applications. Honoring this constraint is imperative for adequate amortization of the hardware development costs. This work addresses these research problems and makes the following contributions: 
\begin{enumerate}
\item 
Based on the observation that computations involving large matrices are pervasive in data analytics and machine learning workloads, we propose the use of digital stochastic circuits for approximate matrix multiplication and develop a high-level abstraction to model the error introduced by this stochastic hardware.
\item
Using this abstraction, we analyze the impact of approximate computation on the gradient descent algorithm and present new techniques by which the stochastic hardware can augment algorithm design and improve execution time.
\item We train deep neural networks for the MNIST handwritten digit classification problem. We observe that networks trained in the presence of hardware noise yield error rates no worse than those trained using precise computations.
\end{enumerate}

\section{Matrix Multiplication using Stochastic Hardware}
\label{sec2} The foundations of stochastic computing circuits can be traced to the work by Poppelbaum~\cite{Pop67} and Gaines~\cite{Gaines69} in the late 1960s, coinciding with the early days of computing revolution. After a fallow period that spanned several decades, there has been a discernible renewal of interest~\cite{alaghi2013survey,miao2013parallel,knag14native,alaghi2014fast} in this rather unconventional method of information processing. In this section, we present the key concepts of stochastic computation and extend them for implementing approximate matrix multiplication. We focus our attention on the multiplication operation for two main reasons: 1) From an application perspective, general matrix multiplication (GEMM) represents the most computationally expensive function within any basic linear algebra subprogram (BLAS Level 3) library implementation, and 2) From a hardware implementation perspective, multipliers consume significantly more resources in terms of area and energy than adders and subtractors. The hardware circuit complexity for a $n$-bit binary tree multiplier is $\mathcal{O}(n^2)$, and $\mathcal{O}(n)$ for a $n$-bit full adder. As discussed next, stochastic circuits significantly reduce the complexity of hardware implementation of certain arithmetic functions, providing an opportunity to achieve a high degree of parallelism and a corresponding improvement in computational performance.

\textbf{Stochastic representation.} Within the stochastic computation framework, a number $x \in [0,1]$ is represented as a $N$-bit long Bernoulli sequence, $\textbf{X} = \{X_1, X_2, ..., X_N\} $ such that the binary random variable $X_i$ takes the value 1 with a probability equal to $x$ i.e. $P(X_i = 1) = x$ \footnote{This can be generalized to $x \in [-1,1]$ using the so-called bipolar stochastic representation~\cite{Gaines69}}.~Each of these $N$ stochastic bits can be generated by comparing $x$ against a random sample drawn independently from $\mathcal{U}[0,1]$ -- assigning $X_i$ to 1 if $x$ is greater than the random sample and to 0, otherwise.The number encoded in a given stochastic bit-sequence can be estimated by counting the average occurrence of 1s in the sequence.

\textbf{Scalar multiplication.} Encoding numbers as probabilities allows for implementation of arithmetic operations such as addition and multiplication using simple digital logic gates at the cost of introducing non-deterministic errors in the computation. Consider two scalar quantities $a$ and $b$ scaled appropriately to lie in $[0,1]$. Let $\textbf{A}, \textbf{B}$ be a $N$-bit long stochastic sequences representing $a$ and $b$, respectively. By definition,
\begin{equation}
P(A_i = 1) = a \text{, and } P(B_i = 1) = b  \text{ }\forall i = {1, 2 , \ldots, N}
\end{equation}
Let $\textbf{C}$ be a stochastic bit-sequence obtained by performing a bit-wise logical AND operation on sequences $\textbf{A}$ and $\textbf{B}$.  It is assumed that the digital hardware implements exact AND gates i.e. AND$(0,0) = 0$, AND$(0,1) = 0$, AND$(1,1) = 1$ with probability 1. Therefore,
\begin{equation}
\begin{aligned}
P(C_i =1) &= P(A_i=1)P(B_i=1) = ab\\
P(C_i =0) &= 1-P(A_i=1)P(B_i=1) = 1-ab
\end{aligned}
\end{equation}
The expected value and the variance of the binary random variable $C_i$ can be expressed as:
\begin{equation}
\mathbb{E}(C_i) = ab, \text{ and }
\mathrm{Var}(C_i) = ab(1-ab)
\end{equation}
The number represented by the Bernoulli sequence $\textbf{C} = \{C_1, C_2, ..., C_N\} $ may be viewed as a random variable obtained by averaging the $N$ independent binary random variables $C_i$. i.e.
\begin{equation}
\label{varScaProd_0}
\begin{aligned}
C &= \frac{1}{N}\sum _i^NC_i \\
\implies \mathbb{E}(C) = ab, \text{ and }
\mathrm{Var}(C) &= \frac{ab(1-ab)}{N} \leq \frac{1}{4N} \forall a,b \in [0,1]
\end{aligned}
\end{equation}
For large $N$, invoking the central limit theorem, multiplication using stochastic number representations produces an unbiased estimator of the product, corrupted with zero-mean Gaussian noise and variance that is inversely proportional to $N$ -- the number of bits used in the stochastic sequence. 
It is possible to extend this analysis to $a,b \in [0,r]$, by sampling the random number used for generating the stochastic bit from $\mathcal{U}[0,r]$. In such a case, the error variance in Eq.~\eqref{varScaProd_0} needs to be modified as:
\begin{equation}
\label{varScaProd}
\mathrm{Var}(C) = \frac{ab(r^2-ab)}{N} \leq \frac{r^2}{4N}
\end{equation}
Note that error variance depends on the values of the numbers being multiplied, and tends to zero as the inputs approach the limits $0$ and/or $r$. 

\textbf{Vector inner product.} The stochastic computation methodology described above can also be applied to vector dot product and matrix multiplication. Consider two vectors $\textbf{a},\textbf{b} \in \mathbb{R}^d$ and define $c_0 = \left\langle \textbf{a},\textbf{b} \right\rangle$, the inner product of vectors $\textbf{a}$ and $\textbf{b}$. We assume that each component of $\textbf{a}$ and $\textbf{b}$  $ \in [0,r]$. $c_0$ can be estimated by generating $N$ stochastic bits for each of the $d$ components of $\textbf{a}$ and $\textbf{b}$, and counting the occurrence of 1 in the bit-wise AND of the $Nd$ bits representing vectors $\textbf{a}$ and $\textbf{b}$.
\begin{equation}
\begin{aligned}
C &= \sum_{j = 1}^d\frac{1}{N}\sum_{i=1}^N C_{j,i}\\
  &= \sum_{j = 1}^d a_jb_j + \sum_{j = 1}^d\mathcal{N}\left(0,\frac{a_jb_j(r^2-a_jb_j)}{N}\right)\\
\end{aligned}
\end{equation}
For uncorrelated stochastic bit-sequences,   
\begin{equation}
\label{varDotProd}
\begin{aligned}
C &= c_0 + \mathcal{N}(0,\sigma^2) \text{, and}\\
\sigma^2 &= \sum_{j = 1}^d\frac{a_jb_j(r^2-a_jb_j)}{N} \leq  \frac{c_0(dr^2-c_0)}{N} \leq \frac{dr^2}{4N}
\end{aligned}
\end{equation}
\vspace{-1em}
\begin{figure}
	\centering
   	\includegraphics[scale=0.6]{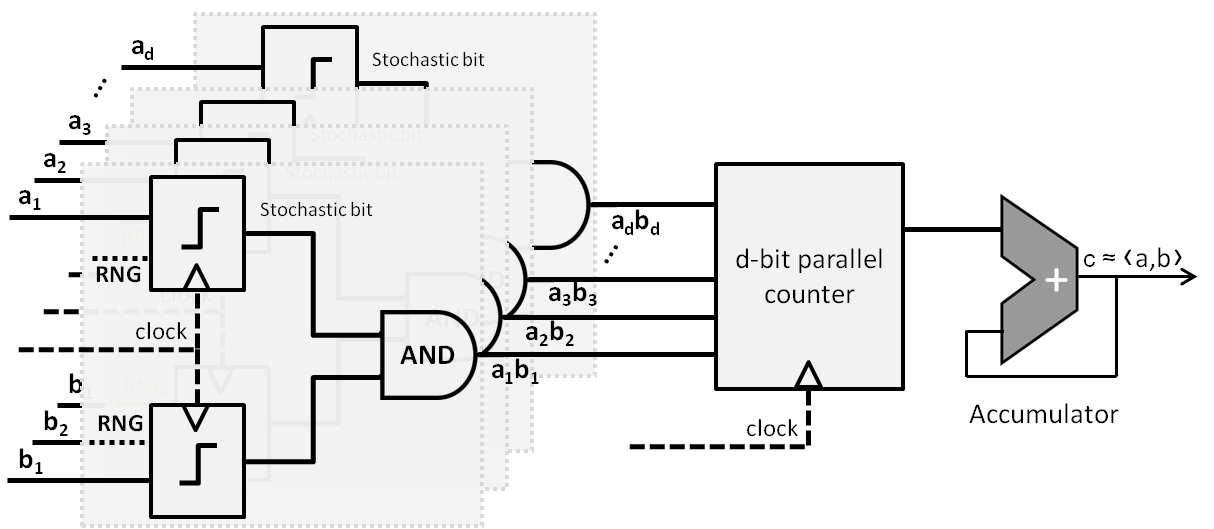}
	\caption{A stochastic multiply-and-accumulate unit. This schematic shows a parallel implementation of the inner product of vectors \textbf{a}, \textbf{b} $\in \mathbb{R}^d$}
	\label{fig:Fig_1}
\vspace{-1.5em}
\end{figure}

Figure~\ref{fig:Fig_1} shows the schematic of a possible hardware implementation of the vector inner product. This design consists of $d$ units, processing the $d$ components of vectors \textbf{a} and \textbf{b} in parallel. In every clock cycle, the $j^{th}$ unit produces a stochastic bit for $a_j$ and $b_j$ by comparing them against random numbers drawn independently from a uniform distribution. The stochastic bits are fed into an AND gate, generating a bit that is set to 1 with probability $a_jb_j$. A $d$-bit parallel counter sums up the 1-bit output of these $d$ units producing an estimate of the inner product. The output of the counter can be averaged over $N$ clock cycles in order to refine the result of the inner product computation. It is preferred to force $N$ to be a power of 2 so that the normalization of the accumulator result by $N$ can be implemented using inexpensive bit shift operations. In this particular design of a multiply-and-accumulate unit, the accuracy of the stochastic computation can be tuned by suitably adjusting $N$ and allows trading off accuracy for improvement in computation time. Interestingly, \textit{this control over the accuracy can be achieved without modifying the underlying hardware},  differentiating stochastic computing circuits from other approximate computing techniques such as low-precision digital circuits and analog circuits \footnote{To some extent, stochastic circuits have certain attributes in common with analog circuits. Both rely on transforming the number representation to enable efficient computations. Analog circuits use physical quantities such as voltages and currents, whereas stochastic circuits work with a more abstract notion of probabilities.}. Compared with the latter, stochastic circuits offer the advantage of seamless compatibility with the state-of-the-art in using standard CMOS logic gates, providing an opportunity for rapid design prototyping and verification using low-cost, commodity FPGAs (for eg.~\cite{alaghi2013stochastic,mansinghka2014building}). 

As compared with low-precision digital circuits, stochastic circuits provide an extremely area-efficient implementation of basic arithmetic functions, but incur significant overheads in terms of generating the stochastic bit-sequences. These overheads include the additional circuitry required for comparators and random number generators, and may potentially limit the degree of parallelism that may be achieved. Addressing these limitations of stochastic computing circuits is an emerging research topic and a diverse set of solutions have recently come forth. For example,~\cite{knag14native} adopts a device-level approach and proposes the use of memristive devices~\cite{strukov2008missing} for stochastic bit-sequence generation. In~\cite{miao2013parallel}, the authors present a parallel stochastic computing architecture that improves the computation speed and accuracy at the cost of increasing the area footprint of the overall system. In an orthogonal approach,~\cite{alaghi2014fast} proposes the use of low-discrepancy quasi-random sequences for generating stochastic bits and improving the speed of stochastic circuits. To augment these efforts, circuit and architecture-level solutions are needed that optimize the place in the system for generating stochastic bit-sequences (near memory, near caches or near the core), hardware interfaces that feed data into the stochastic-bit generator, as well as all the components that might be needed to build a complete ``Stochastic ALU". Given these open research questions regarding the specifics of the stochastic hardware, we defer further discussion to a future report. Nonetheless, the noise model presented above can be used to assess the impact of stochastic computations on the behavior of machine learning algorithms. The findings of this investigation will not only determine the compatibility of stochastic computing for such applications, but also provide valuable insights that can influence hardware design.

\section{Learning in Presence of Hardware-induced Noise}
Machine learning tasks are routinely formulated as an optimization problem  with the aim of finding a set of model parameters that minimizes a well-defined cost function. This optimization problem is typically solved using gradient-based first order techniques. The calculation of the gradient is computationally expensive, and the algorithm may be sped-up by offloading the gradient computation onto a stochastic hardware accelerator. The stochastic hardware returns a noise-corrupted version of the gradient. Given this setting, consider the $k^{th}$ iteration of the noisy batch gradient descent for obtaining $x_*$ that minimizes $f(x): \mathbb{R}^n \rightarrow \mathbb{R}$,
\begin{equation}
	\label{gradDes}
	\begin{aligned}
	x_{k+1} &= x_k - \alpha_k\nabla \tilde{f}\left(x_k\right), \alpha_k>0, \\
	\nabla \tilde{f}(x_k) &= \nabla f(x_k) + G_k
	\end{aligned}
\end{equation}
$\nabla \tilde{f}(x_k)$ is an unbiased estimator of the true gradient $\nabla f(x_k)$ and $G_k$ is a vector representing the error introduced by the stochastic hardware. As shown previously, entries of $G_k$ are i.i.d, satisfying
\begin{equation}
	\label{defG}
	 \mathbb{E}(G_k) = \textbf{0} \text{, and } \mathbb{E}(G_k^TG_k) = \sigma_k^2 \leq \frac{\sigma_0^2}{N_k} 
\end{equation}
where $N_k$ is the length of the stochastic bit sequence used in the $k^{th}$ iteration\footnote{The variance of each component of $G_k$ depends on the input values as shown in Eqs.~\eqref{varScaProd} and~\eqref{varDotProd}. We avoid this additional complexity in the present analysis by bounding the variance from above by a constant that is inversely proportional to $N_k$}. Note that this formulation of gradient descent does not differ appreciably from that of the classical stochastic gradient descent algorithm~\cite{robbins1951,KushnerLin} or gradient descent with noise-corrupted gradient (Proposition 3 in ~~\cite{bertsekas2000}). These proof techniques are directly applicable to the problem in Eq.~\eqref{gradDes} and theoretical guarantees for convergence can be achieved by enforcing strict constraints on the permissible learning rate schedules. The learning rate $\alpha_k$ is required to decrease monotonically and satisfy $\sum \alpha_k =  \infty $ and $\sum \alpha_k^2 < \infty$. However, in a practical machine learning setting, approximate optimization is often sufficient ~\cite{BottouBousquet} and sometimes preferred in order to avoid over-fitting. Given these relaxed constraints, we would like to understand how the hardware-induced error propagates through the successive iterations of gradient descent and more importantly, discover new methods by which the stochastic hardware can improve computational performance. For further analysis of algorithm in Eq.~\eqref{gradDes}, we assume that $f$ is $l$-strongly convex,
\begin{equation}
f(x) - f(y) \leq \nabla f(x)^T(x-y) - \frac{l}{2} \|x - y\|^2, l > 0
\end{equation}
 with Lipschitz continuous gradients
 \begin{equation}
  \| \nabla f(x) - \nabla f(y) \| \leq L\|x - y\|, L > 0
 \end{equation} 
The expected distance of the $k+1^{th}$ iterate from the solution $x_*$, conditioned on the previous iterate, can be expressed as
\begin{equation}
\label{expXK+1}
\begin{aligned}
\mathbb{E}\left(\|x_{k+1}-x_* \|^2| x_k \right)&=  \mathbb{E}\left(\|x_{k}-x_* - \alpha_k\nabla f(x_k) -\alpha_k G_k \|^2| x_k \right) \\
& =\|x_{k}-x_* - \alpha_k\nabla f(x_k)\|^2 + \alpha_k^2\mathbb{E}\left(G_k^TG_k\right)\\
&\leq \|x_{k}-x_* - \alpha_k\nabla f(x_k)\|^2 + \frac{\alpha_k^2\sigma_0^2}{N_k} 
\end{aligned}
\end{equation}
At this point, we can borrow some well-known results from convex optimization of $f(x)$~\cite{boyd2004convex}
\begin{equation}
\label{contMap}
\|x_{k}-x_* - \alpha_k\nabla f(x_k)\|^2 \leq \beta_k^2 \|x_k - x_*\|^2,
\end{equation}
where $\beta_k := 1-\alpha_kl$
\begin{equation}
\label{expXK+1_2}
\begin{aligned}
\mathbb{E}\left[\|x_{k+1}-x_*\|^2\right] &= \mathbb{E}\left[\mathbb{E}\left(\|x_{k+1}-x_* \|^2 | x_k\right)  \right] \\ 
&\leq \beta_k^2 \mathbb{E}\left[\|x_k - x_*\|^2\right] +  \frac{\alpha_k^2\sigma_0^2}{N_k} = \beta_k^2 \mathbb{E}\left[ \mathbb{E}\left(\|x_k - x_*\|^2|x_{k-1}\right)\right] + \frac{\alpha_k^2\sigma_0^2}{N_k} \\
&\leq \beta_k^2\beta_{k-1}^2 \mathbb{E}\left[\|x_{k-1} - x_*\|^2 \right]+ \beta_k^2 \frac{\alpha_{k-1}^2\sigma_0^2}{N_{k-1}} + \frac{\alpha_k^2\sigma_0^2}{N_k} \\
&\leq \left(\prod_{i=1}^{k}\beta_i^{2}\right)\|x_1 - x_*\|^2 + \sigma_0^2\left(\sum_{i=1}^{k}\frac{\alpha_i^2}{N_i}\prod_{j=i+1}^{k}\beta_j^2	\right)\\
\end{aligned}
\end{equation}
If $\alpha_k$ is fixed at the optimal learning rate for batch gradient descent~\cite{boyd2004convex} $\alpha = \frac{2}{l+L}$, and the same bit-sequence length $N_0$ is used in each iteration, Eq.~\eqref{expXK+1_2} can be simplified to
\begin{equation}
\label{case1}
\mathbb{E}\left[\|x_{k+1}-x_*\|^2\right] \leq \beta^{2k} \|x_1 - x_*\|^2 +\alpha^2 \frac{\sigma_0^2}{N_0}\frac{1-\beta^{2k}}{1-\beta^2}
\end{equation}
Since $\beta = 1-\alpha l = \frac{L-l}{L+l} < 1$, the algorithm converges to a random variable with expectation $x_*$ and variance $\sigma_*^2 = \frac{\sigma_0^2}{lLN_0}$. 
\begin{figure}
\vspace{-1.5em}
\centering
\captionsetup[subfigure]{labelformat=empty}
\subfloat[]{
  \centering
  \includegraphics[width=0.5\textwidth]{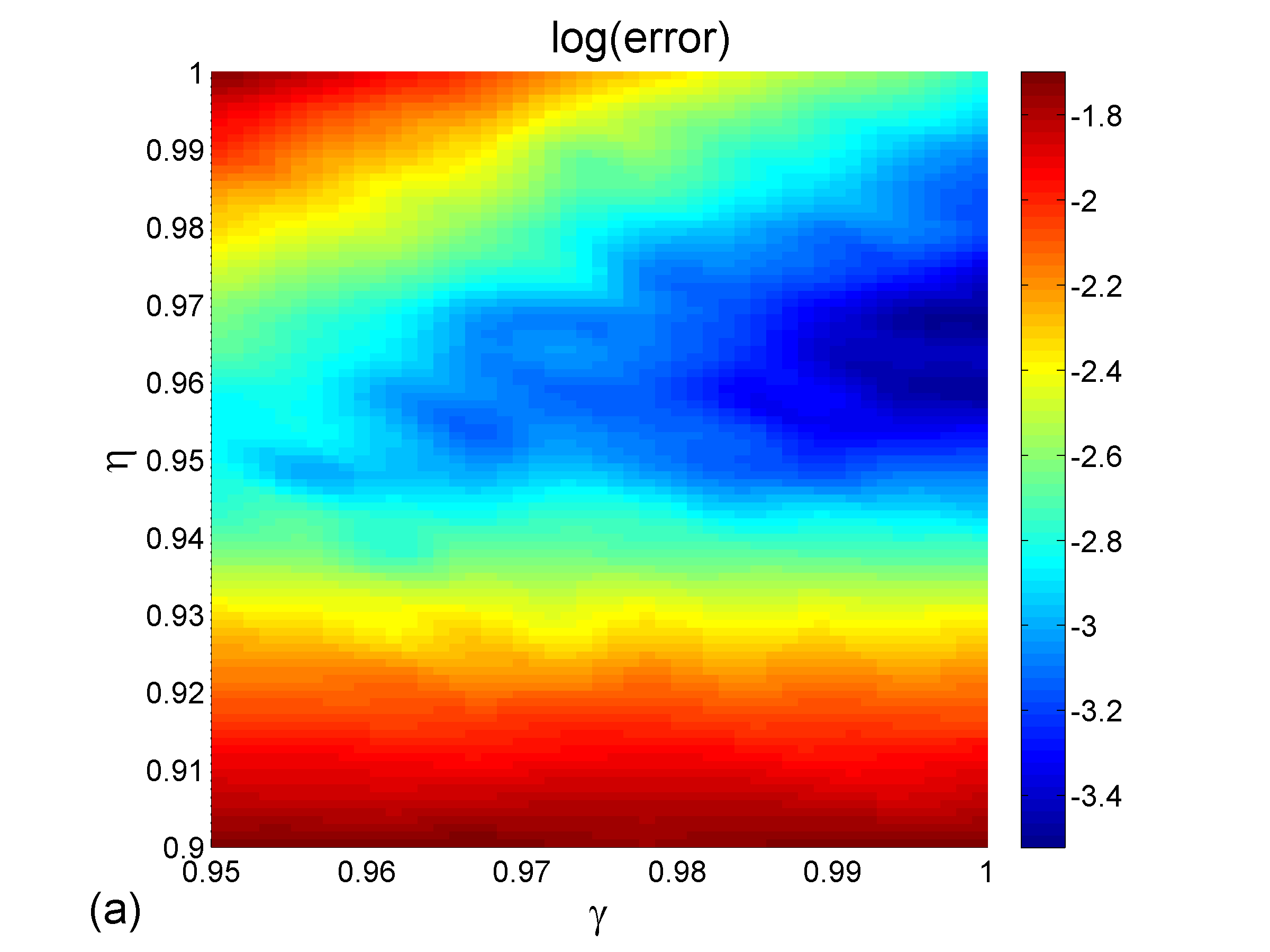}
  \label{fig:logerr}
}
\subfloat[]{
  \centering
  \includegraphics[width=0.5\textwidth]{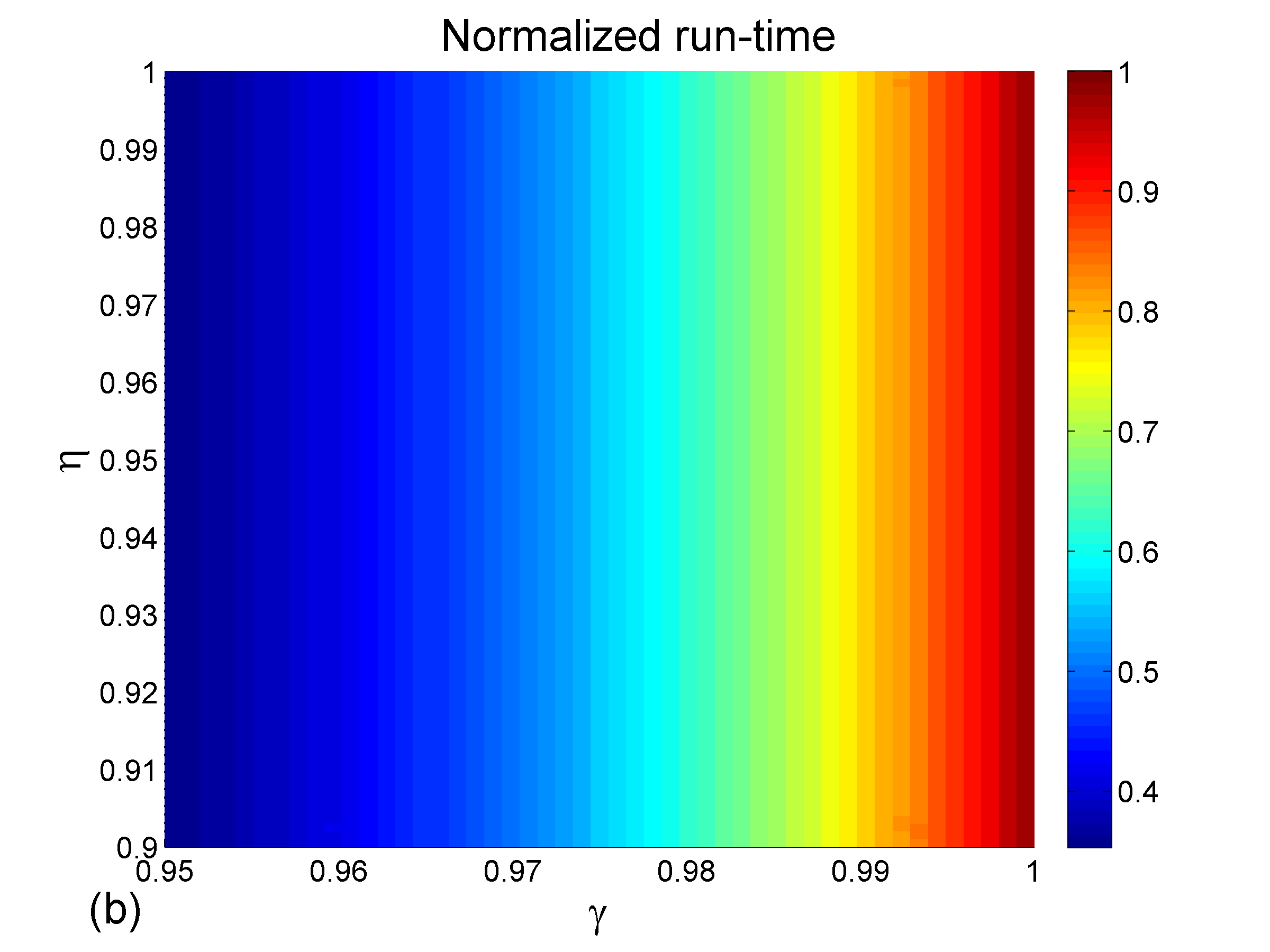}
  \label{fig:runtime}
  }
\vspace{-1.5em}
\caption{(a) Optimization error and (b) normalized total run time for different values of $\gamma$ and $\eta$. The error shown is obtained after averaging over $100$ repetitions. Gradient descent is run for $k=50$ iterations. $\alpha_k = \alpha_0\eta^k$, $N_k = 1 + N_0\gamma^k$, $t_k = T_0N_k$}
\vspace{-1em}
\label{fig:scoobydoobydoo}
\end{figure}

A more intriguing result emerges when we allow $\alpha_k$ decay to exponentially with $k$ i.e. $\alpha_k = \alpha_0\eta^k$, where $0< \eta<1$, and at the same time, let $N_k$ \textit{decrease} with $k$ as $N_k = 1 + N_0\gamma^k$, $0< \gamma\leq 1$. As a consequence of reducing $N_k$, the stochastic hardware is expected to compute the $k^{th}$ gradient faster while producing a less accurate estimate. However, the effect of this additional variance is partially mitigated by the exponentially decaying learning rate, as determined by Eq.~\eqref{expXK+1_2}. For the parallel implementation of stochastic computing circuits as shown in Figure~\ref{fig:Fig_1}, it is reasonable to assume that the computation time for the $k^{th}$ iteration, $t_k$, scales in proportion to $N_k$. To understand the implications of this choice of $\alpha_k$ and $N_k$, we minimize a convex $f(x)$ under different settings of hyperparameters $\eta$ and $\gamma$. The results are shown in Figure~\ref{fig:scoobydoobydoo}. Clearly, there exists a $\gamma < 1$ that yields, statistically, similar optimization error as in the case when $N_k$ is kept constant $\left(\gamma = 1\right)$. Furthermore, for any  $\gamma < 1$ there is an improvement in the total algorithm run-time, arising primarily due to the reduction in the computation time needed for the iterations that are executed using a smaller $N_k$. Note that this improvement occurs in addition to any acceleration by the virtue of offloading the computation onto the stochastic hardware. 

\section{Training Neural Networks using Stochastic Hardware}
As a demonstration of the stochastic computing techniques developed in Section~\ref{sec2}, we consider the problem of training deep neural networks using the back-propagation method. This choice is motivated by the fact that training the deep neural networks is computationally demanding, creating the necessity for efficient hardware acceleration techniques that enable the scalability of the learning algorithm for training large, complex neural network architectures using big training data sets. In addition, the computational complexity and the execution time of the mini-batch stochastic gradient descent (SGD) algorithm typically used for neural network training is dominated by a series of dense GEMM operations in the feed-forward, error back-propagation and weight update calculation steps. Furthermore, the mini-batch SGD is inherently a sequential algorithm -- only limited benefits can be achieved by model-level, data-level parallelism \cite{seide2014} -- and accelerating the dense GEMM operations can immensely improve its computational performance. As a result, mini-batch SGD is particularly well-suited for implementation on stochastic hardware that performs fast, but approximate GEMM. 

We investigate the impact of approximate matrix computations on the classification performance of a deep neural network. We consider the digit classification task on the MNIST dataset. This dataset comprises of $60,000$ training images and $10,000$ test images - each image is $28 \text{x} 28$ pixels containing a digit from $0$ to $9$ and the pixel values are scaled to $[0,1]$. In our experiments the effect of GEMM computation on a stochastic hardware accelerator is modeled by adding a random matrix to the result of a precise computation{\footnote{In the present context, `precise' refers to $64$-bit double precision floating point computation}. Each element of this random matrix is sampled from a Gaussian distribution $\mathcal{N}(0,\sigma^2)$, where $\sigma^2$ is inversely proportional to the length of the stochastic bit-sequence used to represent the numbers. We also modulate the noise variance $\sigma^2$ in accordance with Eqs.~\eqref{varScaProd} and~\eqref{varDotProd}. The following functions are assumed to be computed on the stochastic hardware:
\begin{enumerate}
\item Forward propagation of the input vector across each layer: $Y_{l+1} = W_{l}^{T}X_l + B_l$ 
\item Backward propagation of the error vector across each layer:
\begin{enumerate}
\item[a.]GEMM operation to calculate: $\zeta_l = W_l\delta_{l+1}$  
\item[b.]Hadamard product for evaluating: $\delta_{l} = \zeta_{l}\circ g'\left(Y_{l}\right)$
\end{enumerate}
\item Calculation of the update to the weight matrix: $\Delta W_l = X_l^T\delta_{l+1}$
\end{enumerate}
In the notation used above, $l$ indexes the different layers with $l=1$ corresponding to the input layer, $g$ is the sigmoid activation function, $X_{l+1}:=g\left(Y_{l+1}\right)$ is the input to the $l+1^{th}$ layer. $W_l$ is the weight matrix and $B_l$ is bias vector associated with the layer $l$. We train the neural network using SGD with mini-batch size of 100, and cross-entropy objective function. Momentum ($p=0.5$) is used to speed up the convergence of gradient descent. We adopt an exponentially decreasing learning rate -- scaling it by a factor of $0.99$ after every epoch of training.
\begin{figure}[t]
\vspace{-1.5em}
\centering
\captionsetup[subfigure]{labelformat=empty}
\subfloat[]{
  \centering
 \includegraphics[width=0.4\textwidth]{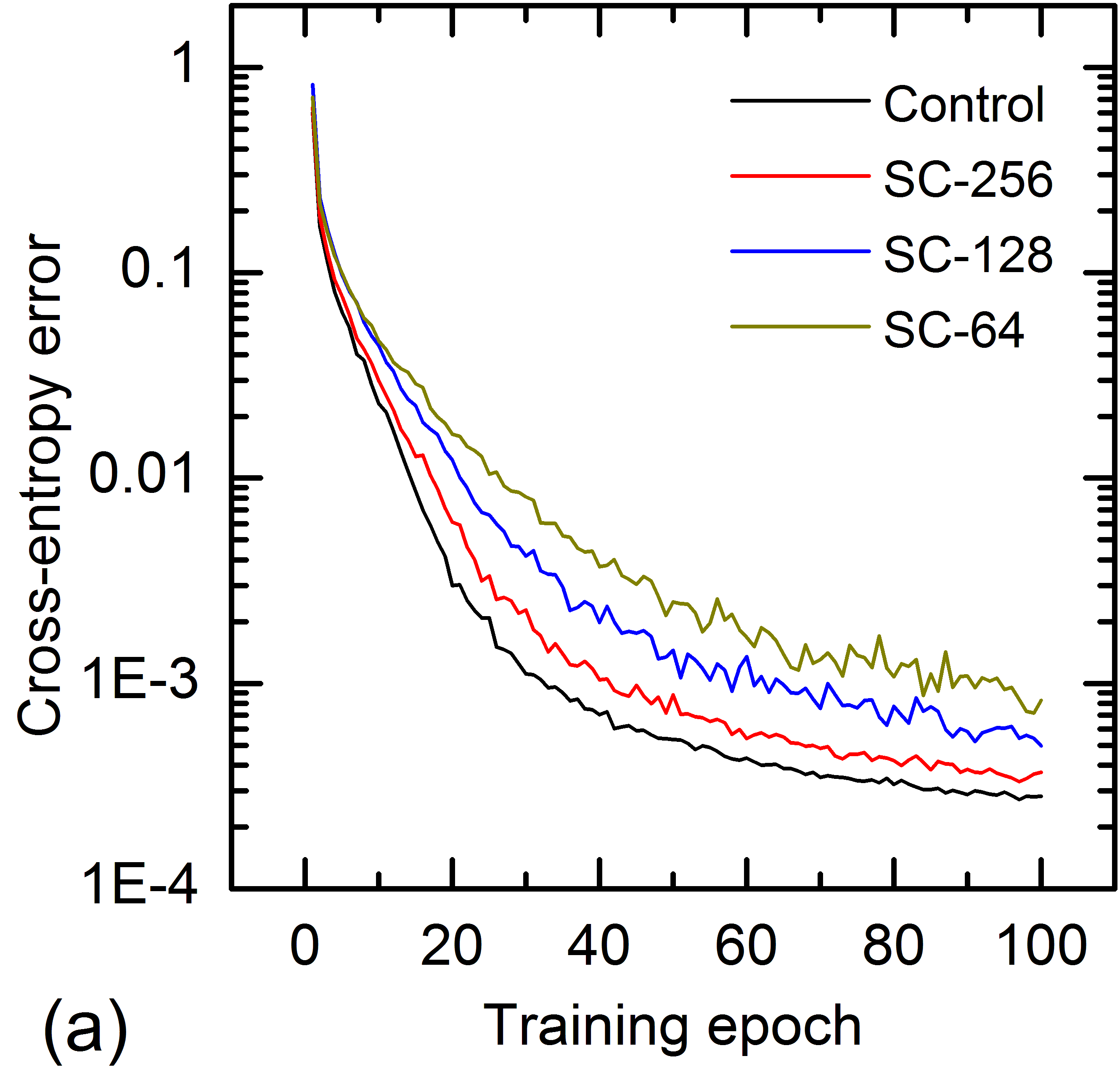}
  \label{fig:noPT_train}
}\quad
\subfloat[]{
  \centering
 \includegraphics[width=0.4\textwidth]{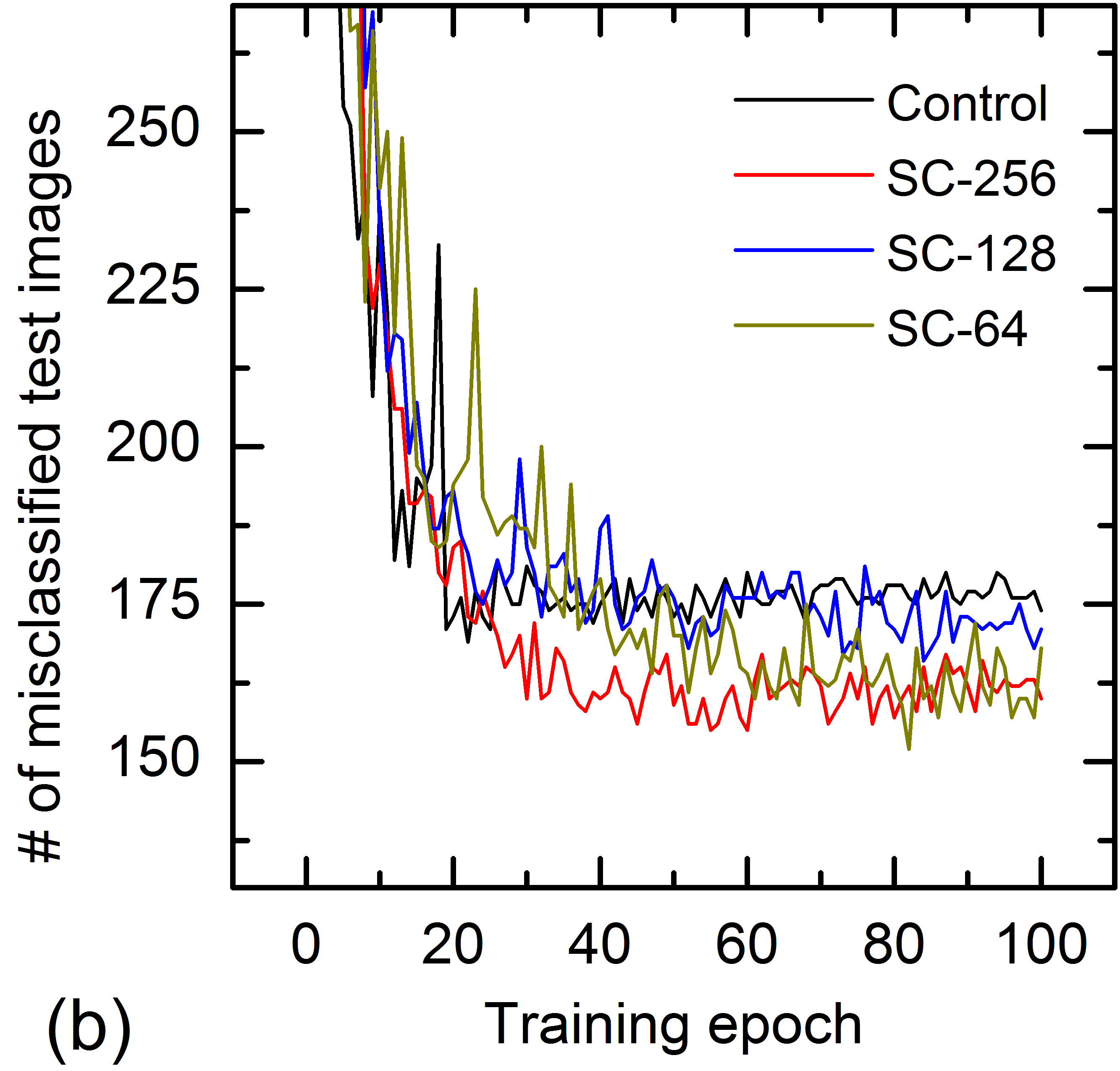}
  \label{fig:noPT_test}
}
\vspace{-1.5em}
\caption{Performance of a 784-400-400-10 neural network under different noise conditions and random initialization of weights. In `SC-$N$', $N$ refers to the length of the bit-sequence used in stochastic computation. The results for a precise computation (`control') are also shown for comparison.}
\vspace{-1em}
\label{fig:noPT}
%\end{figure}
%\begin{figure}
\vspace{-0.0em}
\centering
\captionsetup[subfigure]{labelformat=empty}
\subfloat[]{
  \centering
  \includegraphics[width=0.4\textwidth]{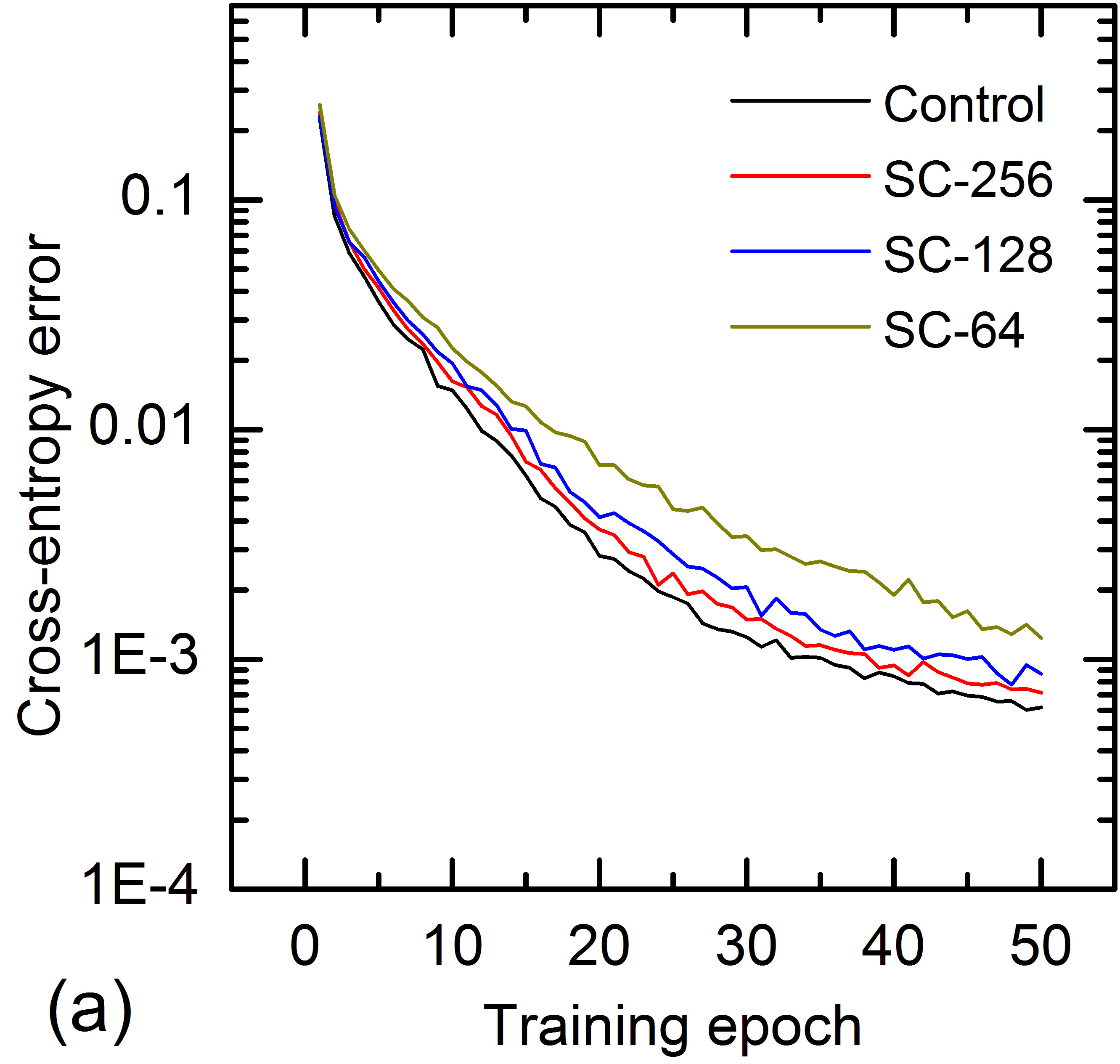}
  \label{fig:PT_train}
}\quad
\subfloat[]{
  \centering
  \includegraphics[width=0.4\textwidth]{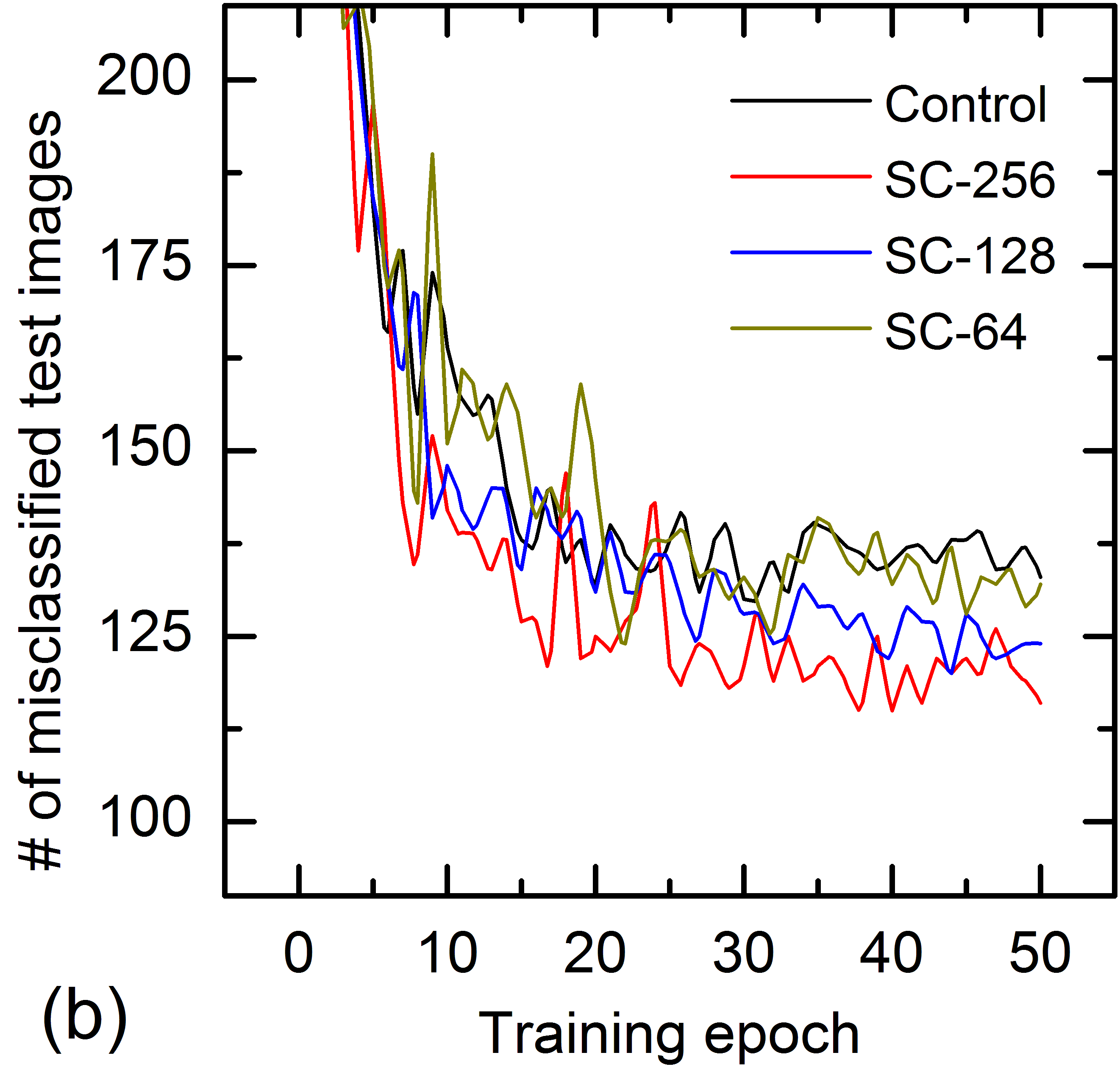}
  \label{fig:PT_test}
  }
\vspace{-1.5em}
\caption{Performance of a 784-400-400-10 neural network for different noise conditions and weight initialization using unsupervised feature learning.}
\vspace{-1em}
\label{fig:PT}
\end{figure}

In the first set of experiments, we construct a neural network with $2$ hidden layers, each containing $400$ units. The weight matrices for each layer are initialized to random values sampled from $\mathcal{N}(0,0.1)$. The bias vectors are initialized to \textbf{0}. The learning rate for the first epoch is set to $2$. Weight decay parameter $\lambda = 0.001$ is used for all the layers. We train this neural network for $100$ epochs under different noise conditions determined by $N$, the length of the stochastic bit-sequence used for number representation. It is important to note that the parameters described above are kept unchanged while training the network in the presence of noise-corrupted computations. As a benchmark for comparison, we train a `control' network using precise computations. Figure~\ref{fig:noPT_train} shows the evolution of the cross-entropy error as the network is trained. We observe a monotonic increase in the cross-entropy training error as $N$ is reduced. However, as shown in Figure~\ref{fig:noPT_test}, networks trained on a stochastic hardware do not suffer any degradation in the classification performance as compared with the control network. On the contrary, there seems to be a slight, but noticeable, improvement in the classification accuracy for training in presence of hardware-induced noise. The control network incorrectly classifies $175$ test images (error = $1.75\%$), where as the network `SC-$256$' yields a test error of $1.62\%$. This result should not come as a surprise, especially when considered under the light of some prior work (for eg.~\cite{bishop95,murray94}) that provide insights into the noise benefits in neural network training. In~\cite{bishop95}, Bishop equates training in presence of input noise to Tikhonov regularization, and similarly in~\cite{murray94}, the addition of weight noise during training is shown to improve the neural network's generalization ability and error-tolerance. It is reasonable to expect that the training on stochastic hardware also serves to weakly regularize the deep network by the virtue of adding noise at the several stages of the backpropagation algorithm.

A variety of different techniques have been proposed to further improve the classification accuracy deep neural networks. These include generative pre-training~\cite{hinton06}, stochastic regularization through dropout ~\cite{hinton12dropout}, dropconnect ~\cite{wan13dropc}, stochastic pooling~\cite{zeiler2013stochastic}, and other architectures for deep networks such as convolutional neural networks. These techniques can be used in conjunction with our approach of training on stochastic hardware. To support this conjecture, we train another set of deep networks in which we initialize the layer weights by performing unsupervised feature learning using stacked sparse autoencoders~\cite{Ngiam2011}. This weight pre-training is also performed on the stochastic hardware. The network is then fine-tuned for $50$ epochs using mini-batch SGD. The results shown in Figure~\ref{fig:PT}, are qualitatively similar to the previous case where the weights were initialized randomly. With weight pre-training, the test error for the `control' network drops to about $1.35\%$. Again, we notice that networks trained on stochastic hardware achieve slightly lower test error than the `control' network. From the results shown in Figures~\ref{fig:noPT_test} and~\ref{fig:PT_test}, no clear trend emerges that would dictate the preference for a particular value of $N$, the length of the stochastic bit-sequence. Rather, we find that there exists a wide range of $N$ which may be used for neural network training without degrading the classification accuracy. We speculate that this range of $N$ depends on the problem at hand as well as the particular choice of network hyperparameters for a given problem. Extending this study to include different datasets, different neural network architectures and performing a more detailed sensitivity analysis will perhaps give further insights. 

\section{Conclusions}
In this paper, we have proposed a framework for acceleration of machine learning algorithms via purpose-built non-deterministic hardware. We have sketched the design of stochastic circuits where numbers are encoded in terms of hardware-instantiated random variables. When standard batch gradient descent procedures for convex optimization of machine learning objective functions run on this stochastic hardware, the noise introduced due to computational errors turns these procedures into variations of stochastic gradient descent that are somewhat different from those commonly considered in the literature. In particular, apart from step-sizes, the stochastic bit-sequence length offers a complimentary knob with which learning algorithms can be orchestrated towards faster convergence. As a proof-of-concept, we have empirically demonstrated that deep learning techniques can be accelerated, with no loss of accuracy, by offloading key back-propagation computations onto stochastic hardware. Extrapolating, we envision the emergence of ``big data" frameworks for machine learning based on relaxed, inexact models of computing running on error-embracing components all across the stack, right down to low-level hardware circuitry.

\small{
\bibliography{nips2014_V4}

\begin{thebibliography}{10}

\bibitem{BottouBousquet}
L{\'e}on Bottou and Olivier Bousquet.
\newblock The tradeoffs of large scale learning.
\newblock In {\em NIPS}, volume~4, page~2, 2007.

\bibitem{KushnerLin}
H.~Kushner and G.G. Lin.
\newblock {\em Stochastic Approximation and Recursive Algorithms and
  applications}.
\newblock Springer, 2003.

\bibitem{Mahoney}
M.~Mahoney.
\newblock Randomized algorithms for matrices and data.
\newblock In {\em Foundations and Trends in Machine Learning}, 2011.

\bibitem{Jordan}
V.~Chandrasekaran and M.~I. Jordan.
\newblock Computational and statistical tradeoffs via convex relaxation.
\newblock In {\em Proceedings of the National Academy of Sciences}, 2013.

\bibitem{Pop67}
WJ~Poppelbaum, C~Afuso, and JW~Esch.
\newblock Stochastic computing elements and systems.
\newblock In {\em Proceedings of the November 14-16, 1967, fall joint computer
  conference}, pages 635--644. ACM, 1967.

\bibitem{Gaines69}
BR~Gaines.
\newblock Stochastic computing systems.
\newblock In {\em Advances in information systems science}, pages 37--172.
  Springer, 1969.

\bibitem{alaghi2013survey}
Armin Alaghi and John~P Hayes.
\newblock Survey of stochastic computing.
\newblock {\em ACM Transactions on Embedded computing systems (TECS)},
  12(2s):92, 2013.

\bibitem{miao2013parallel}
Lifeng Miao and Chaitali Chakrabarti.
\newblock A parallel stochastic computing system with improved accuracy.
\newblock In {\em Signal Processing Systems (SiPS), 2013 IEEE Workshop on},
  pages 195--200. IEEE, 2013.

\bibitem{knag14native}
P~Knag, W~Lu, and Z~Zhang.
\newblock A native stochastic computing architecture enabled by memristors.
\newblock {\em Nanotechnology, IEEE Transactions on}, 13(2):283--293, 2014.

\bibitem{alaghi2014fast}
Armin Alaghi and John~P Hayes.
\newblock Fast and accurate computation using stochastic circuits.
\newblock In {\em Proceedings of the conference on Design, Automation \& Test
  in Europe}, page~76. European Design and Automation Association, 2014.

\bibitem{alaghi2013stochastic}
Armin Alaghi, Cheng Li, and John~P Hayes.
\newblock Stochastic circuits for real-time image-processing applications.
\newblock In {\em Proceedings of the 50th Annual Design Automation Conference},
  page 136. ACM, 2013.

\bibitem{mansinghka2014building}
Vikash Mansinghka and Eric Jonas.
\newblock Building fast bayesian computing machines out of intentionally
  stochastic, digital parts.
\newblock {\em arXiv preprint arXiv:1402.4914}, 2014.

\bibitem{strukov2008missing}
Dmitri~B Strukov, Gregory~S Snider, Duncan~R Stewart, and R~Stanley Williams.
\newblock The missing memristor found.
\newblock {\em Nature}, 453(7191):80--83, 2008.

\bibitem{robbins1951}
Herbert Robbins and Sutton Monro.
\newblock A stochastic approximation method.
\newblock {\em The annals of mathematical statistics}, pages 400--407, 1951.

\bibitem{bertsekas2000}
Dimitri~P Bertsekas and John~N Tsitsiklis.
\newblock Gradient convergence in gradient methods with errors.
\newblock {\em SIAM Journal on Optimization}, 10(3):627--642, 2000.

\bibitem{boyd2004convex}
Stephen~P Boyd and Lieven Vandenberghe.
\newblock {\em Convex optimization}.
\newblock Cambridge university press, 2004.

\bibitem{seide2014}
Frank Seide, Hao Fu, Jasha Droppo, Gang Li, and Dong Yu.
\newblock On parallelizability of stochastic gradient descent for speech dnns.
\newblock In {\em Proceedings of the 39th International Conference on
  Acoustics, Speech and Signal Processing (ICASSP-14)}, pages 235--239, 2014.

\bibitem{bishop95}
Chris~M Bishop.
\newblock Training with noise is equivalent to tikhonov regularization.
\newblock {\em Neural computation}, 7(1):108--116, 1995.

\bibitem{murray94}
Alan~F Murray and Peter~J Edwards.
\newblock Enhanced mlp performance and fault tolerance resulting from synaptic
  weight noise during training.
\newblock {\em Neural Networks, IEEE Transactions on}, 5(5):792--802, 1994.

\bibitem{hinton06}
Geoffrey~E Hinton, Simon Osindero, and Yee-Whye Teh.
\newblock A fast learning algorithm for deep belief nets.
\newblock {\em Neural computation}, 18(7):1527--1554, 2006.

\bibitem{hinton12dropout}
Geoffrey~E Hinton, Nitish Srivastava, Alex Krizhevsky, Ilya Sutskever, and
  Ruslan~R Salakhutdinov.
\newblock Improving neural networks by preventing co-adaptation of feature
  detectors.
\newblock {\em arXiv preprint arXiv:1207.0580}, 2012.

\bibitem{wan13dropc}
Li~Wan, Matthew Zeiler, Sixin Zhang, Yann~L Cun, and Rob Fergus.
\newblock Regularization of neural networks using dropconnect.
\newblock In {\em Proceedings of the 30th International Conference on Machine
  Learning (ICML-13)}, pages 1058--1066, 2013.

\bibitem{zeiler2013stochastic}
Matthew~D Zeiler and Rob Fergus.
\newblock Stochastic pooling for regularization of deep convolutional neural
  networks.
\newblock {\em arXiv preprint arXiv:1301.3557}, 2013.

\bibitem{Ngiam2011}
Jiquan Ngiam, Adam Coates, Ahbik Lahiri, Bobby Prochnow, Quoc~V Le, and
  Andrew~Y Ng.
\newblock On optimization methods for deep learning.
\newblock In {\em Proceedings of the 28th International Conference on Machine
  Learning (ICML-11)}, pages 265--272, 2011.

\end{thebibliography}
}
\end{document}